\documentclass[12pt,oneside]{amsart} 
\usepackage[verbose,tmargin=3cm,bmargin=3cm,lmargin=2cm,rmargin=2cm]{geometry}
\usepackage{fancyhdr}

\usepackage{lscape} 

\usepackage{array}
\usepackage{amsmath}
\usepackage{amssymb}
\usepackage{amsfonts}

\usepackage{tabularx}
\usepackage{booktabs}
\usepackage{multicol}
\usepackage{multirow}

\usepackage[font={small},labelfont=bf]{caption}

\usepackage{natbib}

\usepackage[pdftex,x11names]{xcolor}

\usepackage{url} 

\usepackage{subfig}
\usepackage{graphicx}
\usepackage{placeins} 

\usepackage[singlespacing]{setspace}

\usepackage[T1]{fontenc}
\usepackage[default,osfigures,scale=0.95]{opensans}

\title{Lost in Space: Geolocation in Event Data}
\vspace{\baselineskip}

\author[Lee]{Sophie J. Lee}
\address{Sophie J. Lee: Department of Political Science}
\curraddr{Duke University}
\email{sophie.lee@duke.edu}

\author[Liu]{Howard Liu}
\address{Howard Liu: Department of Political Science}
\curraddr{Duke University}
\email{hao.liu@duke.edu}

\author[Ward]{Michael D. Ward}
\address{Michael D. Ward: Department of Political Science}
\curraddr{Duke University}
\email{michael.d.ward@duke.edu}

\date{\today}

\thanks{We appreciate the comments of Kyle Beardsley, Andrew Hall, Jan Kleinnijenhuis, Sayan Mukherjee, Jonathan Nagler,
Molly Roberts, Colin Rundel, David Siegel, Brandon Stewart, Joshua Tucker, members of Wardlab at Duke University, 
and the SMaPP lab at New York University. This research was partially supported by the National Science Foundation Award 1259266. \\ Corresponding Authors: ophie J. Lee and Howard Liu.}

\newcolumntype{d}{D{.}{.}{-1}}
\bibpunct[:]{(}{)}{;}{a}{}{;}

\begin{document}

\title{Lost in Space: Geolocation in Event Data}

\begin{abstract} 
\noindent Extracting the ``correct'' location information from text data, i.e., determining the place of event, has long been a goal for automated text processing. To approximate human-like coding schema, we introduce a supervised machine learning algorithm that classifies each location word to be either correct or incorrect. We use news articles collected from around the world (Integrated Crisis Early Warning System [ICEWS] data and Open Event Data Alliance [OEDA] data) to test our algorithm that consists of two stages. In the feature selection stage, we extract contextual information from texts, namely, the N-gram patterns for location words, the frequency of mention, and the context of the sentences containing location words. In the classification stage, we use three classifiers to estimate the model parameters in the training set and then to predict whether a location word in the test set news articles is the place of the event. The validation results show that our algorithm improves the accuracy rate of the current geolocation methods of dictionary approach by as much as 25\%.
\vspace{1cm}

\end{abstract} 

\maketitle

\section{Introduction}

Many quantitative studies of conflict rely on event data. Recently, these studies have also retreated from the country-year framework and have focused on disaggregating the event flows both in terms of space and time. Disaggregating temporality\textemdash even to the daily level\textemdash is a straightforward task. But figuring out precisely \textit{where} an event actually occurred is a difficult and uncertain task that has been perplexing for most contemporary event data efforts \citep{icews:2015:data}, PHOENIX \citep{oeda:2016}, SCAD \citep{salehyan:2015}, ACLED \citep{raleigh:etal:2010}.

At the same time there is widespread interest in  disentangling investigations from the country-year framework.  The country-year---as an observational framework---has a longstanding tradition in political science. Indeed  two of the three most cited articles in the \textit{American Political Science Review} focus on the country-year \citep{beck:katz:1995,fearon:laitin:2003}. As recently as 2009, collections have  focused on disaggregation of the country-year in conflict studies \citep{cederman:gleditsch:2009}. More broadly, many  focus on hierarchical approaches that simultaneously include subnational, national, and even international aspects. Efforts at the World Bank and other international organizations \citep{frank:martinez:2014,eaton:etal:2011} have emphasized this deeper dive into the political and economic landscape. Much of this deeper dive is coming from organizations and governments in terms of their reporting on demographic, economic, financial, and health data that are subnational.  

Most data in the conflict realm comes from non-offical sources. For many that means some form of data collected from historical and journalistic sources. This need is often filled by event data, which are typically collected on a daily basis, and can be aggregated temporally to the level required by the analysis.  Event data can also be aggregated to the geographical region that is appropriate.  
Given the increasing demands for event data, the scientific community has recently devoted significant efforts to automate the data collection process. Having humans read and code a large set of archive documents sometimes limits reproducibility, and hence hinders scientific research. It is also expensive and limits the currency of the data. Further, ensuring inter-coder reliability is challenging, especially over global events that span decades.

Several efforts utilize machine-coding to collect event information and determine event features automatically. Projects such as the Integrated Crisis Early Warning System (ICEWS) \citep{icews:2015:data,icews:2015:aggregations,obrien:2010} and the Open Source Event Data Alliance \citeyear{oeda:2016} are two prominent examples. A good overview on event data in political science is found in \citet{schrodt:yonamine:2013}. These automated event data allow researchers to observe and extract information on politically relevant events around the world in near real-time.

Despite the apparent advantages of automated data collection, the machine-coding ontologies for event data require further research 
\citep{grimmer:stewart:2013,lucas:etal:2015}.  NSF currently sponsors a multidisciplinary project to look into formulating the generation of event 
data.
\footnote{Modernizing Political Event Data for Big Data Social Science Research, Patrick T. Brandt (PI, EPPS), Vito D'Orazio (Senior Personnel, EPPS), Jennifer S. Holmes (Co-PI, EPPS), Latifur R. Khan (Co-PI, ECS), Vincent Ng (Co-PI, ECS), National Science Foundation, RIDIR, 
\$1,497,358, September 2015---August 2018.}
At the same time, IARPA is reportedly looking to fund an event data challenge that could lead to new ways of collecting and analyzing event data. This attention by funding agencies illustrates that not all issues in this research domain are yet resolved. Outstanding issues include machine translation of texts in foreign languages (wherein great progress is being made in both Chinese and Arabic), duplicate reports from multiple sources, and the relatively low accuracy in determining the event location \citep{dorazio:etal:2014,schrodt:2015}. While all of these are important, in this article, we focus on the sole issue of geolocation in event data.

For human coders, locating events by reading a news article may be time-intensive, but straightforward. This is not the case for machine-coding: many news articles contain multiple location names, such as the location of the journalist writing the story, the birthplace of a person being interviewed, or the place of a similar event that occurred several decades ago; at times, human names are identical to geographic names; and location names are transliterated into English in a variety of potentially confusing ways. All these sources of \textit{noise} in the data increase the difficulty in automatically locating events. A good algorithm should read texts like human coders and code only the correct event locations.

We treat the geolocation task as a classification problem where each location word is predicted to be correct or incorrect. With the goal of developing an algorithm to discern correct event locations automatically, we extract the contextual information of location words (the N-gram patterns for location words, the frequency of mention, and the context of the sentences containing location words) from the training set. We check the accuracy against a hand-coded set of ground truth data on locations. To do so, estimated parameters from the training set were used to predict the event locations in the test set (for which we also know the ground truth), using three classification methods: artificial neural networks with back propagation, support vector machine (SVM), and random forests. Our supervised machine learning language model codes locations correctly at an accuracy rate close to 90\% when the texts contain a single correct location per article.\footnote{Upon publication a repository of our project will be available at: \\texttt{https:github.com/(author ID hidden)/LostInSpace}.} 
 
Our approach is fully automated, but does require some hand coding of a small number of stories to contextualize the coders for specific countries. While the process described in this article is generalizable, we selected data from ICEWS \citep{icews:2015:data} and OEDA.\footnote{For the OEDA project, which is still in early development, see: \texttt{http:phoenixdata.org}.} We began with an investigation of 250 protests in China (CAMEO code 14: protest), 
but to anneal the generalizability of our approach, we added (and coded) 250 violent events in the Democratic Republic of the DRC (CAMEO code 19: fight) drawn from the ICEWS data, and 250 violent events drawn from the PHEONIX data on Syria (OEDA, CAMEO code 19: fight).\footnote{While there are twenty action verbs in the CAMEO ontology, verbs such as ``appeal,'' ``consult,'' or ``yield'' do not yield as many events in the data as ``protest,'' ``assault,'' or ``fight'' do. Also, the data of such events are not in the settings of interstate or intrastate conflicts. Hence we selected ``protest'' and ``fight'' data as test cases. Our method, however, is applicable to all the other CAMEO event types.} Each of these cases presents difficult problems for automated geolocation.

\section{Automated Geolocation}

The task of determining event locations involves three steps, each non-trivial. In the first step, known as \textit{named entity recognition} in Computer Science and Computational Linguistics \citep{dorazio:etal:2014,cardie:wilkerson:2008,guerini:etal:2008,arguello:etal:2008,nadeau:sekine:2007}, all location names are identified and extracted from an appropriately preprocessed text. This step is a prerequisite for the other steps because to determine the location of an event in a news article, capturing the exhaustive list of location names is required. Next comes entity disambiguation/resolution, which involves identifying the actual location of the recognized name string\citep{cucerzan:2007,dorazio:etal:2014,bunescu:pasca:2006}. Once this is accomplished, it is possible to extract the ontologically defined meaning from the text in terms of who does what to whom, and when and where via 
CAMEO,\footnote{For more information on CAMEO ontology, please see: \texttt{http:eventdata.parusanalytics.com/data.dir/cameo.html}.} 
PETRARCH,\footnote{See \texttt{https:github.com/openeventdata/petrarch}.} or some other coding framework \citep{schrodt:2006}. Lastly, the disambiguated location names are evaluated to determine whether they represent the event-occurring location. While this step requires the completion of the two preceding steps, the enormous, extant body of work that has built up over the past decades of machine-coding of events has made the first two steps more manageable. 

Named entity recognition in the context of geolocation involves determining which words in the given sentences are location names. In principle, the task of capturing location names from texts can be done easily by using a dictionary. In practice, however, developing a dictionary that is sufficiently comprehensive for such a task may be challenging. To begin, the geographic boundary of the texts being analyzed may be unclear, given that the domain of many event data is the entire world. Further, because conflict events often spread to new and rural places, texts may include location names not defined in the gazetteer. Still further, a location name may be written in multiple forms, requiring the dictionary to comprise every variant for each location. This commonly occurs when a foreign location name is transliterated into another language such as English. For instance, the transliterations \textit{dei ez-zor}, \textit{Deir-al-Zour}, \textit{Dayr al-Zawr}, and \textit{dei ez-Zour} all refer to the same province in Syria. Without specific dictionaries and correspondences, this is difficult to determine automatically. 

Further complicating matters, news articles often use nearby landmarks to indicate the location, in lieu of using the official names. The 2014 Ukrainian revolution, for example, was often described as having taken place at Mariinsky Park, rather than in Kiev. The same is true for the so-called Martyr Square (Tahrir Square) and its role as a site of protests during the Egyptian revolution of 2011. We encountered this problem in our data set as well. For example, ``U.N. attack helicopters whirred overhead as armoured personnel carriers ploughed through forests in Virunga National Park [a park in North Kivu province] in Democratic Republic of Congo \ldots''.\footnote{ICEWS story ID: \texttt{4590482}, DRC data}

The dictionary approach can be complemented by the part-of-speech (POS) tagging method that grammatically parses sentences in the text and classifies each word into various categories such as persons, organizations, and location names. The technique can identify landmarks (for example, Mriinsky Park, Martir Square, or Virunda National Park) as locations even if they were not pre-defined in the dictionary. Currently, there are a number of open source systems available for named entity recognition. Typical software programs for this task include the Stanford Named Entity Recognizer (part of Stanford NLP), Apache Open NLP algorithm, and MIT Information Extraction (\textit{MITIE}), developed by MIT's Lincoln Laboratory. 
The parsing stream for the OEDA pipeline, for example, combines POS tagging and the dictionary approach.

\begin{figure}
\includegraphics[scale=.5,angle=-90]{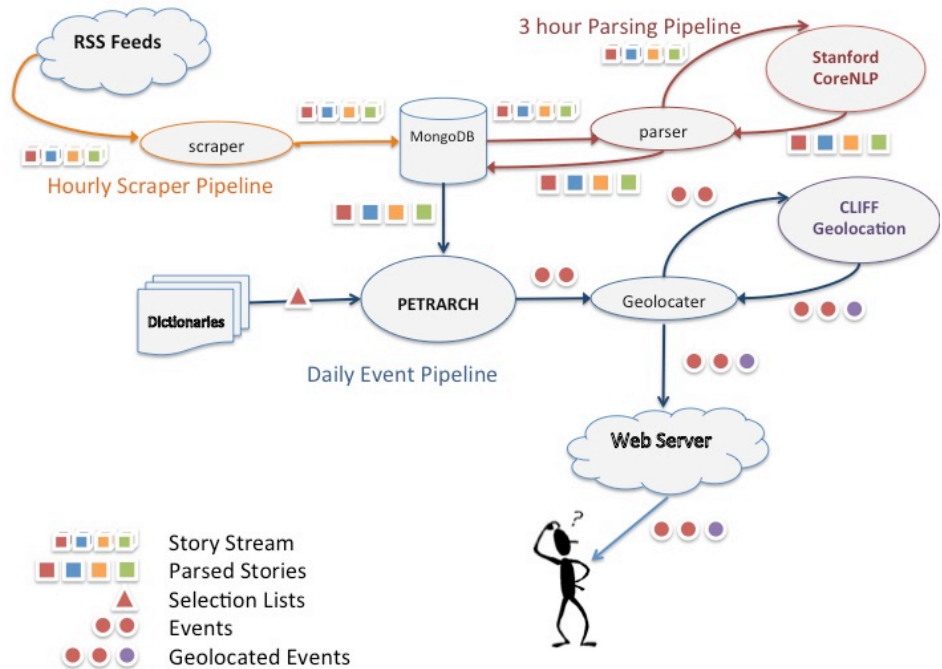}
\caption{\label{fig:pipeline} Schematic Representation of OEDA Pipeline, circa June 2016}
\end{figure}

As shown in Figure~\ref{fig:pipeline}, in OEDA the parser calls the Stanford CoreNLP, which returns to the Mongo database parsed sentences with parts of speech tagged. These parsed stories then are coded via the PETRARCH ontology. Only then is geolocation undertaken, using calls to the CLIFF geolocation software. While this method extracts an extensive list of location words, coding all country-relevant location words as the correct event location introduces a different problem.

Entity disambiguation is a nascent field of research aiming at determining the true location of the referent location word. \citep{han:etal:2011, rao:etal:2013, bunescu:pasca:2006, cucerzan:2007}. For instance, ``Durham'' in the sentence,``the group moved to the intersection of Duke and Chapel Hill streets near the Durham Police Department headquarters'' (Jul. 21st, 2016. CBS News Carolina), would most likely to refer to a city in North Carolina, U.S., while the same word in the sentence ``Paul Collingwood commits for another year to Durham'' (Jul. 26th, 2016. AFP) would most likely to refer to a city in England. 

Different approaches exist for assigning location name strings to the referent location words. The co-occurences of location names, i.e. which location words frequently appear together in the corpus, could be modeled and used to link the name strings and the true locations \citep{han:etal:2011}. Similarly, the \textit{Mordecai} algorithm links the extracted location names to the locations defined in a gazetteer, by adopting \textit{Word2vec} model \citep{mikolov:etal:2013} that calculates the co-occurrences of the words and quantifies the contexts in which specific words appear.\footnote{Mordecai is described at \texttt{https:github.com/openeventdata/mordecai}. } A well-trained corpus archive should be able to show words such as ``Duke'' and ``Chapel Hill'' appear commonly with ``Durham'' when ``Durham'' is the city name in North Carolina. While the disambiguation task is not simple, well-defined dictionaries may suffice these techniques when processing news articles that have clearly defined country bounds. But for projects that contain new and unknown sub-national location names, building an extensive dictionary is a daunting task, especially if the locales are not named in English.

Finally, geolocating events (identifying the location of the event described in a document) is an objective for many scholars, particularly those who intend to collect and build original databases from text corpora, be they news articles, congressional records, campaign speeches, party constitutions, or twitter feeds. While automating this task will aid many, the research avenue in this topic is still under development. One of the most commonly used methods is building location dictionaries and capturing location names. For instance, the principal investigators of the Project Civil Strife (PCS) data used three location dictionaries\textemdash``cities,'' ``regions/provinces,'' and ``others''\textemdash and coded all captured location names as the place of event \citep{shellman:2008}. This approach, however, also includes irrelevant places as event locations. In our China data, for example, a total of 614 location words were captured from 250 news articles but only half of them (314 correct location words) are actual event locations. Given that a substantial number of location words  are incorrect event locations, the automated event data community needs a better coding scheme that can reduce the error rates. Many have noted that this problem is yet to be solved. We discuss the remaining challenges in detail in the ensuing section.  

\section{Challenges in Geolocating Events}

Selecting the correct location word among all captured locations is a difficult problem. In the data we examined, nine out of ten news articles contain multiple locations. Some of these are specious locations, indicating for example the location of news agencies, the location of a similar, often previous event, or the current location of a reporter. The occurrence of multiple location names not only introduces noise into the data, but also escalates the difficulty in automating the task of geolocation. Noting this difficulty, \citet{schrodt:yonamine:2012} states that ``the main challenge is to empirically determine which place name should be assigned to the specific event, especially when multiple events and location names occur in a single article'' (page 19). 

Multiple approaches have been devised. One approach, adopted by ICEWS, is to code the location name that is the nearest to the action verb (identified by the TABARI coder\footnote{For the TABARI coder see: \texttt{http:eventdata.parusanalytics.com/software.dir/tabari.html}.}) in the text. Under such a scheme, a location word that is distant from the action verb is automatically discarded. Although the rationale for the algorithm sounds intuitive, errors frequently occur because action verbs are not always adjacent to the names of the event location. 

The current OEDA data deployed uses a java-based web service named CLIFF.\footnote{CLIFF documentation: \texttt{https:github.com/openeventdata/phoenix\_pipeline} and \texttt{http:cliff.mediameter.org} } This approach selects the most likely place as the ``focus'' location of the article, based on the frequencies of mentions and the order of appearance \citep{dignazio:etal:2014}. 

Notably, both ICEWS and OEDA assume that one correct event location exists per article. Yet, that assumption does not always hold. Over 30\% of Syrian stories we examined contain multiple true locations, and in China and the DRC data, these numbers are about one-third to one-half this amount. For the other 175 countries in the world, these ratios, to our knowledge, are not yet known, but we can assume that the ratio is greater than zero. As the article in Table \ref{tab3:tn} demonstrates, single stories frequently contain multiple true location names. By assuming that a single location word exists per article as ``the most appropriate location'' \citep{lautenschlager:etal:2016}, such a coding rule misses many true event locations, thereby hindering the accuracy of the coded location names. Moreover, a researcher who analyzes an event data under these approaches may misinterpret the data, for example, concluding that protests in China are concentrated in the capital, Beijing. However, even when there is a single location, the OEDA and ICEWS geolocation still will make many mistakes.

On the other hand, coding all location names in a text as the correct locations, as the PCS project has done, reduces false negatives but increases false positives. Hence, the optimal approach would determine which set of location words in the article is more likely to be the correct ones, in addition to relaxing the assumption that only a single location word represents each news article.  
  
\section{Categorizing Multiple Locations}
In examining articles with multiple locations, we observed four mutually exclusive types of location words: 1) event-relevant and event-occurring, 2) event-irrelevant and event-occurring, 3) event-relevant and non-event-occurring, and 4) event-irrelevant and non-event-occurring. We define event-relevant locations as those locations that are part of the main description of the event of interest, i.e. all locations that are key to the narrative of the event of interest. Event-occuring locations refer to all locations where events occurred regardless of whether the event is the event of interest. Thus, the first category, \textit{event-relevant and event-occurring}, refers to the locations where events occurred while the occurred events are within the scope of interest. We aim to detect this type of location words as the correct ones. 

\begin{table}[t]    
  		\caption{Location Word Type 2 (\textcolor{red}{Incorrect}) Vs. Type 1 (\textcolor{blue}{Correct});  ICEWS story ID: \texttt{17929606}, DRC data; and  OEDA story ID: \texttt{2054559 v0.1.0}, Syria data }
			\label{tab:error2}
			\begin{minipage}{\textwidth} 
			\begin{center}
			\begin{tabular}{p{0.08\hsize} p{0.90\hsize}} \toprule[0.02in]
			\multicolumn{2}{c}{ \textcolor{red}{event-\textbf{ir}relevant and event-occurring} Vs. \textcolor{blue}{event-relevant and event-occurring} }\\			\midrule
		  \multirow{1}{*}{Eg. 1} &	The Syrian Observatory for Human Rights monitor said that 15 civilians, among them 11 children, were killed in the attack on the Bab al-Nayrab neighborhood in the south of \textcolor{blue}{Aleppo}.\ldots Meanwhile, a ceasefire has been agreed in the town of Daraya, allowing 700 rebel gunmen safe passage to the northern province of \textcolor{red}{Idlib} and allowing 4,000 women and children to escape to shelters outside the town.\\
      \midrule
      Eg. 2 & 1.\ldots The government denounces a named refugee camp near \textcolor{blue}{Goma} that was attacked by M23 soldiers\ldots\\
 &6. Prime Minister Ponyo addressing an opening session of a seminar on agricultural sector in \textcolor{red}{Kinshasa} today.\ldots \\
					
			\bottomrule[0.02in]
			\end{tabular}
			\end{center}
			\end{minipage}
\end{table}

Regarding the second category, a small portion of articles contain \textit{event-irrelevant and event-occurring} location words in our data. Such could occur when the raw texts contain news summaries of events that are not of interest. Table \ref{tab:error2} shows examples of articles that contain both event-irrelevant and event-occurring (Idlib and Kinshasa) as well as the event-relevant and event-occurring (Aleppo and Goma) location words. The first example is from an article that describes a rebel attack event involving 15 civilian causalities and then describes a ceasefire agreement, an event not of interest. Sometimes, news articles contain summaries of completely unrelated events as in the second article, which consists of six reports that summarize events that occurred in Syria on that day.

The third type is \textit{event-irrelevant and non-event-occurring} locations. Some of the most commonly observed event-irrelevant and non-event-occurring location names refer to the location of the news agencies and spokespersons. For example, Beijing  in the first article in Table \ref{tab:error3} indicates where the story was being written, but coded as the actual event location in ICEWS. We suspect that this is because the location word that refers to the reporting location is the closest location name from the action verb (strike), and hence was mistakenly coded as the event location. The true event location in this article is Guangdong.

  \begin{table}[b]		
			\caption{Location Word Type 3  (\textcolor{red}{Incorrect}) Vs. Type 1 (\textcolor{blue}{Correct});
			ICEWS story ID: \texttt{18141520}, China data;  ICEWS story ID: \texttt{10672170}, DRC data.}
			\label{tab:error3}
			\begin{minipage}{\textwidth} 
			\begin{center}
			\begin{tabular}{p{0.08\hsize} p{0.90\hsize}} \toprule[0.02in] 
			
			\multicolumn{2}{c}{	\textcolor{red}{event-\textbf{ir}relevant and \textbf{non}-event-occurring} Vs. \textcolor{blue}{event-relevant and event-occurring} }\\
      \midrule
          Eg. 1 &  \textcolor{red}{Beijing}, Nov 19, 2011 (AFP) - More than 7,000 workers went on strike at a southern Chinese factory\ldots Dozens of workers were injured on Thursday as police tried to break the strikers' blockade of the main road in the factory town near Dongguan in \textcolor{blue}{Guangdong} province...    \\
\midrule          
    Eg. 2 & The clashes, which started at Luozi and to Seke Banza [towns in \textcolor{blue}{Bas-Congo} Province]\ldots 
    Speaking to reporters in \textcolor{red}{Kinshasa}, the parliamentarian said that the clashes had left at least 100 people dead and many more nursing serious injuries since last Friday.\\
      \midrule
		Eg. 3 &	The protesters gathered outside the office of Southern Weekly in Guangzhou, capital of southern \textcolor{blue}{Guangdong} province, on Monday calling for media freedom, a taboo subject in the country, holding banners and chanting slogans.\ldots A foreign ministry spokesperson in \textcolor{red}{Beijing} is reported to have said: ``There is no so-called news censorship in China.''	\\

						\bottomrule[0.02in]
			\end{tabular}
			\end{center}
			\end{minipage}
\end{table}

Reporting locations often appear in the first line of the article. Discarding the reporting locations can alleviate the problem to a certain extent, but the problem persists because reporting locations are frequently embedded in the middle of texts, as demonstrated in the second and the third articles in Table \ref{tab:error3}. Also, the event-irrelevant and non-event-occurring location words are embedded for other reasons, such as referring to the birthplace of someone being interviewed, as in: `` `we can mix in any society,' said Amar Aldoura from Damascus.''\footnote{OEDA story ID: \texttt{1424875 v0.2.0} }

The \textit{event-relevant and non-event-occurring} locations complicate matters even more. Journalists often provide the background of the event being described. They may recite locations of the stronghold of a rebel group, the province name to which victims fled, or the place where the perpetrators of incidents are being trialed. The articles in Table \ref{tab:error4} are examples of stories containing both event-relevant but non-event-occurring (Orientale and Damascus) and event-relevant and event-occurring locations (North Kivu and Daraa). In the first article, Orientale is a place to which the rebel leader was heading, so the location word is mentioned as part of the description of the rebel attack. But the actual attack was in North Kivu. The writer of the second article mentioned Damascus to describe a goal that the rebel group wishes to achieve. The actual attack was in Daraa.
 
\begin{table}[!htbp]    
			\caption{Location Word Type 4  (\textcolor{red}{Incorrect}) Vs. Type 1 (\textcolor{blue}{Correct}); OEDA story ID: \texttt{1160025 v0.2.0}, Syria data; ICEWS story ID: \texttt{18922379}, DRC data.}
			\label{tab:error4}
			\begin{minipage}{\textwidth} 
			\begin{center}
			  \begin{tabular}{p{0.08\hsize} p{0.90\hsize}} \toprule[0.02in]
			\multicolumn{2}{c}{	\textcolor{red}{event-relevant and \textbf{non}-event-occurring} Vs. \textcolor{blue}{event-relevant and event-occurring} }\\
			\midrule
		Eg. 1	&  The mutineer general, Bosco Ntaganda and his men who have been on the run for several days now, exchanged ``heavy'' gunfire with the army in the night of 7 May and is heading toward the Virunga National Park [a park in  \textcolor{red}{Orientale} province].\ldots ``After four hours of exchange of heavy gunfire at Kibumba,'' a locality at the border of the Nyiaragongo and Ruthshuru territories, in the volatile province of  \textcolor{blue}{North Kivu}, ``we were supported by shots from heavy weapons,'' stated the captain.\\
    \midrule
    Eg. 2 & Elsewhere in Syria, 51 rebel factions operating in the southern province of Daraa announced a campaign to wrest control of areas of Daraa city [capital of \textcolor{blue}{Daraa} governorate] from the government.\ldots SANA reported that an attack by terrorists had been thwarted, with fighter jets pounding rebel targets in surrounding villages. If successful, it would grant the rebels a rear supply base to mount operations on \textcolor{red}{Damascus}\ldots\\
						\bottomrule[0.02in]
			\end{tabular}
			\end{center}
			\end{minipage}
\end{table}

To tackle the issue of creating the exhaustive list of location words that should be considered, we combine existing named-entity recognition, POS tagging, and entity resolution (matching location strings referenced in a gazetteer) techniques. Furthermore, based on the assessment of the types of multiple locations, we have come to the conclusion that each location word should be evaluated and determined whether it is an event-relevant and event-occuring location. For determining the boolean status (true event location or not) of each captured location word in the exhaustive list, we adopt a classification approach, which we discuss more in the next section. A sophisticated algorithm would distinguish the correct locations from the incorrect ones by filtering out the \textit{event-irrelevant} locations, as well as \textit{non-event-occurring} ones.

\section{Classification Algorithms}

For classifying location words either as correct or incorrect, various machine learning techniques could be used, such as the following: Neural Networks \citep{muller:reinhardt:2012, mehrotra:etal:1997,
cheng:titterington:1994},\footnote{See \citet{zhang:zhou:2006, ng:etal:1997} for examples of neural networks
applications in text analysis.} SVM
\citep{cristianini:shawetaylor:2000,  vapnik:1995,
cortes:vapnik:1995},\footnote{Examples of SVM applications in text
analysis can be found in \citet{minhas:etal:2015,tong:koller:2001,joachims:1998}.} random forests
\citep{liaw:wiener:2002, breiman:2001},\footnote{See \citet{fette:etal:2007} for examples. } AdaBoost
\citep{freund:schapire:1997}, K-nearest neighbors (K-NN) \citep{dasarthy:1990},
and naive Bayes \citep{zhang:2004, murphy:2006}. Of these, we employ three
classifiers\textemdash artificial neural networks, SVM, random forests\textemdash and compare the performance of each.

The artificial neural network models the relationship between a set of input
signals, the  {desired feature from the texts}, and an output signal\textemdash {whether a location word refers to the event location}\textemdash using concepts borrowed from our understanding of how a human brain processes information from sensory dendrites through neurons while allowing the impulse to be weighted according to its relative importance (\textit{model parameters}).
Although the algorithm is notoriously slow, the artificial neural networks have more flexibility in terms of structures and parameters compared to other classifiers \citep{lantz:2013, zurada:1992}. But a potential downside of neural network model is that its prediction performances usually relies on a considerable amount of training data.

SVM refers to support vector machines. These were initially introduced for
solving two-group classification problems where the data are mapped into a
higher dimensional input space and construct an optimal separating hyperplane
\citep{vapnik:1998, vapnik:1995}. This approach is often viewed as superior to
other machine learning algorithms, including neural networks, because the
quadratic programming guarantees reaching the global optimum, which often leads
to the larger overall classification accuracy \citep{li:2003,  vapnik:1998,
joachims:1998, maroco:etal:2011}. Furthermore, SVM models are typically less
prone to over-fitting \citep{mukherjee:etal:1997}. SVMs also provide a
computationally efficient way to achieve a reasonably accurate model
\citep{amami:etal:2015, li:2003}.

Finally, the random forests (or decision tree forest) model, championed by Leo
Breiman (\citeyear{breiman:2001}) and Adele Cutler \citep{breiman:cutler:2007},
combines the principle of bagging with random feature selection to add
complexity to the decision tree models. After the ensemble of classification
regression trees (hence the name forest) is generated, the model combines these
trees' predictions. Because the ensemble uses only a small, random portion of the full feature set, the model can 
handle large data sets wherein the high dimensionality  may cause other
models to fail. Despite the difficulty in interpreting the results, it is an all-purpose approach that performs well on most problems  \citep{lantz:2013}. 

These classifiers boast two primary strengths. The first is that they do not
require pre-specifying the type of relationship between the covariates and the
response variable. They are powerful information extraction tools that can
capture underlying relationships not explained by known structures
\citep{jones:linder:2015, lantz:2013, gunther:fritsch:2010, beck:etal:2000}.
Second, these models achieve accurate prediction rates given large enough input
sizes \citep{maroco:etal:2011, hsieh:etal:2011, beck:etal:2000}. 
\citet{lantz:2013} suggest that these algorithms are the most accurate state-of-
the-art approaches, and make few assumptions about the data. 

Some scholars oppose the use of the machine learning in fields that require substantive
interpretations of the parameters \citep{demarchi:etal:2004} because they are
``difficult to interpret'' \citep{lantz:2013}. Despite such
pitfalls, the prediction performance of these models make them 
attractive for geolocation. Whether they produce interpretable results 
can not be determined \textit{a priori}.

\section{Building Dictionaries}
Before classification can begin, correctly formatted text data with desirable features is required. We developed four types of dictionaries in order to preprocess the text data. First and foremost, a {location} dictionary for each country was compiled. The initial location lists were imported from Geonames, Wikipedia, and Google map. These dictionaries contained province names (standardized province and  governorate names) and sub-province names (city, village and town names, spelling variations of both province and sub-province names, and frequently used famous location names) as two separate columns. To ensure that our location dictionary was as comprehensive as possible, we used an iterative process to build it. After the initial location dictionary was built, we went back to the text data and parsed sentences using MITIE. The parts of speech elements classified as location words were then sent to the Genomes API and the returned entity pairs that were not already in, but should have been in the dictionary, were added.

We also developed dictionaries for {actors}. While we imported the actor lists from ICEWS and OEDA data and manually modified them depending on the salient actors in each country, the entire process can be done manually. Without the prior knowledge about the events of the data at hand, one may resort to sentence parsers and build dictionaries iteratively as we did for the location dictionary.  

For the {relevant words} dictionaries, we first imported action verb lists from the CAMEO ontology, on top of which our data sets were built. The verbs for the protest data included words such as ``rally,'' ``demonstrate,'' and ``march.'' For the fight data, the verb list included ``air-strike,'' ``bomb,'' and ``shoot.'' For both dictionaries, we then added key nouns that capture the context of the location sentence such as ``bloodshed'' and ``casualty.'' Likewise, a dictionary including irrelevant words, such as ``report'' and ``interview'', was compiled. As in the process of building the other dictionaries, the relevant words dictionary does not have to depend on any existing ontology but we chose to adopt the pre-existing framework of CAMEO because those action verbs were used to collect the news articles in our data in the first place.

Finally, dictionaries containing {generic words} that are not data specific were compiled. The lists included names of news agencies (for example, AFP, AP, and CNN), directional words (southern, southeastern, \ldots), the names of months (january, jan, february, feb, \ldots), and days (monday, mon, \dots). All of these dictionaries were used as part of preprocessing.

\section{Preprocessing the Texts}

The literature on text analysis describes a few common preprocessing steps. Following \citet{dorazio:etal:2014}, we first removed punctuation and special characters from the text data that contain sentences with location words. We next converted all sentences to lower letters to avoid confusion in recognizing word patterns. We then removed stop words in English\citep{shellman:2008,monroe:etal:2008}. The stop words list was imported from the Stanford NLP Group, but we modified it to exclude prepositions related to locations, such as ``in'', ``at'', and ``from''.  Next, we performed stemming \citep{grimmer:stewart:2013}, using Porter Stemmer \citep{porter:1980}.\footnote{We were careful to preserve important information. For instance, Porter Stemmer removes `ing' at the end of each word, so we converted some province names such as ``liaon'' back to ``liaoning''.}

In addition to the tasks performed prior to most text analysis projects, we also performed two additional text treatment tasks that are critical in our algorithm: 1) \textit{homogenization} of location words and 2) \textit{generalization} of texts using the dictionaries described above. As with many other text analysis projects, the accuracy of our algorithm depends highly on the quality of the pre-treatment process. 

The homogenization step is important because the use of location names in news articles is not always consistent with respect to the spelling of location names, particularly of those in non-English speaking countries. In addition to the transliteration issue, the different conventions of stating locations also complicate the process. For example, news articles by local agencies cite only city names while those by national or international agencies often indicate only province names. Accordingly, we used the location dictionaries to standardize these variations across data. For the sentences that contain only the lower level location words, we converted the administrative division names (city) to higher level ones (province/governorate) with the prefix of ``sub-''. For instance, ``Fataki, Orientale" would be converted to ``sub-orientale, orientale.'' In building the dictionaries, we used the administrative division at the time of the news reports. For instance, city A in year 2005, the year of the event, may be in
province B, but in province C in year 2016. In such a case, we used province B as the correct province. 

As the last step in the preprocessing stage, we generalized the news texts using the aforementioned dictionaries of actors, relevant and irrelevant words (action verbs, key nouns, irrelevant verbs and nouns), numbers, dates and news agencies.\footnote{The entries in these dictionaries were stemmed.} The purpose of this step is to ensure that the algorithm would recognize the following two N-grams as identical\footnote{Location names are generalized in the algorithm after a specific location name is collected. Hence, they should not be generalized during the preprocessing stage.}: ``33 people in Beijing'' and ``2000 people in New York''.\footnote{If these two phrases are generalized in terms of numerals and location names, they become ``numeral people in location''.} More generalized sentence patterns are desirable because the approach aims to match patterns of phrases and sentences from different news articles. An example of a preprocessed text looks like the right side of Table \ref{tab3:tn}.

\begin{table}[!ht]
			\centering
			\caption{Example of Raw and preprocessed Text;ICEWS story ID: \texttt{25149588}, China data}\label{tab3:tn}
			\begin{minipage}{\textwidth} 
		\begin{tabular}{p{0.5\hsize} p{0.5\hsize}} \toprule
				Raw text & 
				Treated text \\
				 \midrule
				BEIJING, Dec 4 (AFP) -- More than 500 people protested outside government offices after a man died under suspicious circumstances while in police custody, a human rights group and relatives said Thursday. It is one of two such cases -- in northern Shandong province and southeastern Fujian province\ldots earlier this year the beating death of a man in a police detention hall in southern Guangdong sparked widespread criticism.\ldots

				&
				 NUMERAL ACTOR ACTION-VERB outsid ACTOR man ACTION-VERB suspici circumst ACTOR custodi ACTOR  relat said DAY.  NUMERAL case DIRECTIONAL shandong ADMIN DIRECTIONAL fujian ADMIN\ldots earlier  DATE  ACTION-VERB ACTION-VERB   man  ACTOR detent hall  DIRECTIONAL LOCATION spark widespread criticism.\ldots   \\
				\bottomrule
			\end{tabular}
	\end{minipage}
	
		\end{table}

\section{Implementation of Automated Classifiers}

To mimic the way human coders would parse sentences and retrieve the relevant information, we trained the machine to learn the collocation patterns of the correct and incorrect location words and then to predict the correctness of a new set of location words based on the collocation patterns of those new words. This approach of storing patterns and solving problems known as case-based reasoning is a common paradigm in automated reasoning and machine learning 
in which a reasoner solves a new problem by using a similar problem that has already been solved \citep{kolodner:1992,demantaras:1997}.

\begin{figure}
\includegraphics[width=\textwidth]{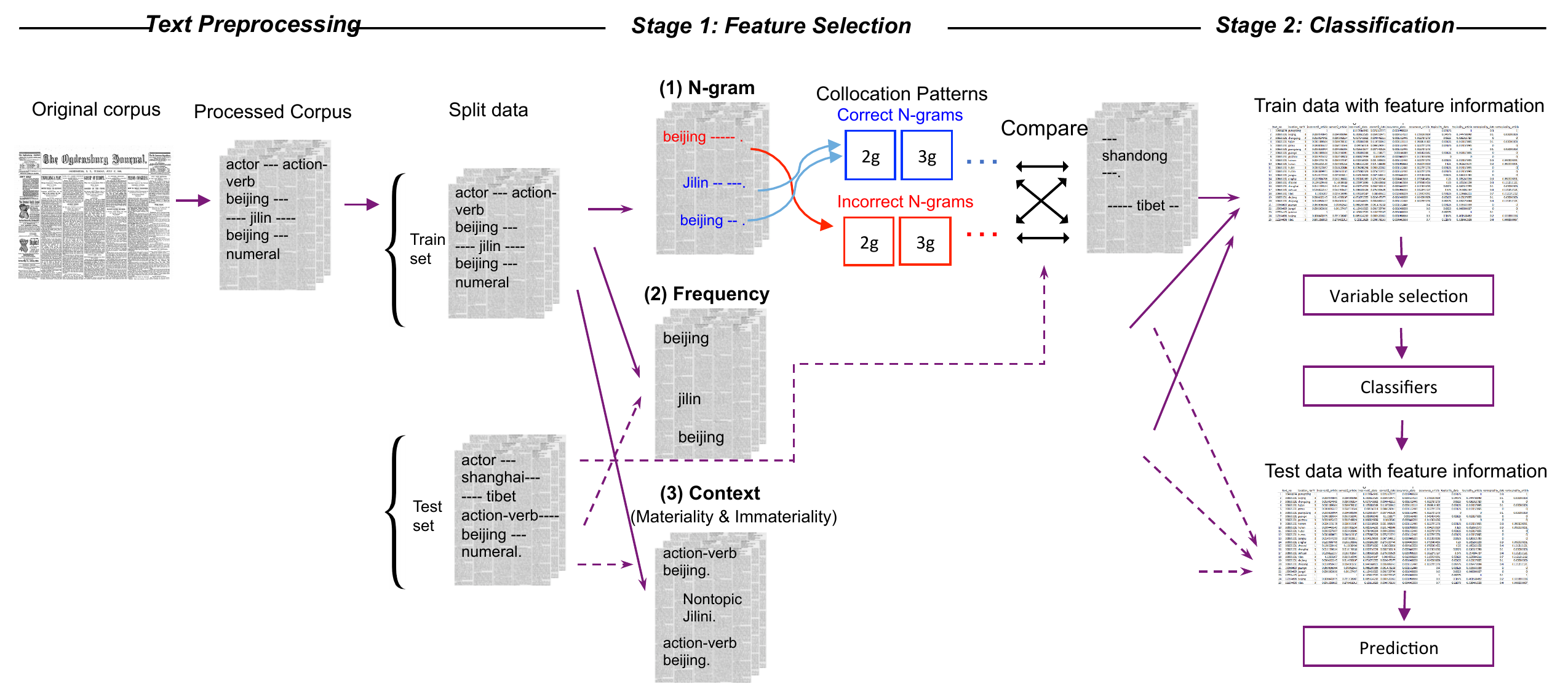}
\vspace{1cm}
\caption{Flow Chart} \label{fig:flowchart}
\end{figure}

The implementation of our algorithm involves two stages: feature selection and model estimation. These stages require pre-treatment of the text data as illustrated in Figure \ref{fig:flowchart}. To describe the feature selection stage in detail, we take examples from the China data, which consist of 250 news articles on protest towards the government from 2001 to 2014.\footnote{We removed duplicate reports.} On average, each article contains about 398 words before the preprocessing treatment and 284 after the treatment. For each validation score (of the nine results that are averaged and presented in Table \ref{table:validation}), we randomly divided the treated articles into training  and  test
sets.

What information then do we feed the machine to develop classifiers? Our goal is to differentiate event-relevant words from irrelevant ones and event-occurring words from non-event-occurring ones. Therefore, we select variables that can provide information about ``event-relevance'' and ``event-occurrence.'' Specifically, 1)  {N-gram} collocation patterns, 2)  {frequency} of location words, and 3)  {context} of the sentences that contain the location word are extracted from the news articles.

An N-gram is a sequence of N words. Collections of N-grams are known to provide valuable information about each word in a phrase, taking into account  the complexity and long distance dependencies of languages.\footnote{For more information on N-gram, see Ch. 4 in \citet{jurafsky:martin:2009}.} In a sentence ``Factory workers protested", an N-gram of order 2 (or bigram) is a two-word-sequence of words (for example ``factory workers", and ``workers protested") while an N-gram of order 3 (or trigram) is a three-word-sequence of words (such as ``factory workers protested"). Given that the collocation patterns in which the event-occurring location words appear differ from those of the non-event-occurring collocations, the N-gram patterns are able to provide the contextual information of event-occurence to our classifiers. We thus compare the frequencies of each N-gram in correct and incorrect corpus and compute the relative frequencies, respectively.

Some examples of N-grams collected from one of the training sets are shown in Table \ref{tab:bigrams}. From the raw text on the left in  \ref{tab3:tn}, ``LOCATION MONTH''\textemdash the bigram for Beijing\textemdash will be stored in the incorrect bi-gram corpus while ``DIRECTIONAL LOCATION'', ``LOCATION ADMIN'', \textemdash the bigrams of Shandong\textemdash and ``DIRECTIONAL LOCATION', ``LOCATION ADMIN''\textemdash the bigrams of Fujian\textemdash will be stored in the correct bigram corpus. Some examples in the bi-grams (N-grams of N=2) of correct location words from the training set in the China data include ``LOCATION ADMIN" (frequency: 17), ``LOCATION ACTION-VERB" (frequency: 15), and ``outsid LOCATION" (frequency: 15). These bi-grams were extracted from sentences such as the following: ``The \textit{Guizhou provincial} government deployed thousands of police",\footnote{Story ID: \texttt{35682875}, ICEWS China data}  ``Workers at IBM Systems Technology Company (ISTC) in \textit{Shenzhen are protesting} since March",\footnote{Story ID: \texttt{32977476}, ICEWS China data} and ``500 villagers had been protesting \textit{outside the Qingdao} naval base".\footnote{Story ID: \texttt{32852391}, ICEWS China data} Two of the most frequent bi-grams in the incorrect corpus are ``LOCATION MONTH", and ``LOCATION ACTOR". They are from phrases such as ``\textit{BEIJING, Dec} 3, 2007 (AFP)",\footnote{Story ID: \texttt{22997344}, ICEWS China data}  ``\textit{Shandong farmers} protested\ldots ".\footnote{Story ID: \texttt{21984369}, ICEWS China data} Table \ref{tab:bigrams} shows the top 10 most frequent bi-grams for both correct and incorrect location words in one of the training sets.

\begin{table}[!ht]
\begin{center}
   \caption{Examples of Collocation Patterns (n=2)}\label{tab:bigrams}
    \begin{tabular}{c|lr|lr}
   & \textbf{Correct} & Freq& \textbf{Incorrect} & Freq\\
        \hline
1   &   of SUB-LOCATION  &	  69 & LOCATION MONTH	&27\\
2	&	of LOCATION	&33& LOCATION ACTOR	&20\\
3	&	in LOCATION &30& to LOCATION &19\\
4	&	DIRECTIONAL LOCATION &24& LOCATION SOURCE &	18\\
5	& in SUB-LOCATION &	24& from LOCATION	&16\\
6	&	LOCATION ADMIN &	17& by SUB-LOCATION &15\\
7	&	LOCATION and	&17& link LOCATION	&14\\
8	& SUB-LOCATION near	&16& NONACTION-VERB SUB-LOCATION	&12\\
9	&	LOCATION ACTION-VERB  &	15& to LOCATION	&11\\
10	&	outsid LOCATION & 15& near LOCATION	&10\\
    \end{tabular}

    \end{center}
\end{table}

Then we compare the captured collocation patterns, consisted of location words and their neighboring words, to the correct and incorrect N-gram lists. In the texts in Table \ref{tab:workingexample}, for instance, ``heilongjiang'' and ``beijing'' would be captured. While creating covariates for ``heilongjiang'', the N-gram collocation patterns, such as ``of heilingjiang'', ``at sub-heilongjiang'', and ``in subheilongjiang'', are converted to ``of LOCATION'', ``at sub-LOCATION'', and ``in sub-LOCATION''. For ``beijing'', the N-gram collocation patterns such as ``to beijing'', ``beijing therefor'', and ``in beijing''  would be converted to ``to LOCATION'', ``LOCATION therefor'', and ``in LOCATION''. These generalized N-gram patterns are compared to the correct and incorrect pattern lists (compiled from the training set) that looks like the list in Table \ref{tab:bigrams}. Then the N-gram pattern feature is converted to numeric values in two ways. The first N-gram variables compute the ratio the collocation patterns comparing both the stored correct and incorrect pattern lists. The other N-gram variables reflect how many of these collected patterns can be matched to the most frequent patterns in each list.\footnote{We used the top 50\% of N-gram patterns in terms of frequencies. For the size of our data, the most frequent correct and inccorrect N-gram lists included 10 to 20 collocation patterns.} 

The second type of variables, the frequencies of location words, provide the information about the relevance of a particular location word. Assuming that the news articles in the data are well-sorted and contain articles mostly pertinent to the research interest, the set of location words that are mentioned several times should have higher chances of being correct, compared to the ones with low frequencies \citep{dignazio:etal:2014}.  

\begin{table}[!ht]
      \centering
			\caption{Illustrative Example, ICEWS story ID: \texttt{25149588}, China data with Province name added by the author.}\label{tab:workingexample}
			\begin{minipage}{\textwidth} 
  		\begin{tabular}{p{0.5\hsize} p{0.5\hsize}} \toprule
    Raw text & 				Treated text \\
				 \midrule
				About 500 angry textile workers blocked a railway line in northeastern China on Monday demanding unpaid wages and unemployment pay from the government, said railway employees who saw the protest.  There were four to five hundred of them blocking the railway. They stood on the railway, but were later dispersed," said a man who worked at a railway station in Jiamusi [town in \textcolor{blue}{Heilongjiang}] the northeastern province of \textcolor{blue}{Heilongjiang}.
"The train to \textcolor{red}{Beijing} was therefore delayed five to six minutes in leaving," he said. The protest was similar to one in December when more than 2,000 workers from the same bankrupt textile plant blocked a rail line and cut traffic on an airport highway, accusing company officials of embezzling their social security payments. An employee at Jiamusi [town in \textcolor{blue}{Heilongjiang}]'s main train station said city and railway police went to persuade them to leave and protesters dispersed after about 20 minutes, he said. 
The plight of disgruntled workers laid off from bloated state-owned firms, like those in Jiamusi [town in \textcolor{blue}{Heilongjiang}], is getting top billing at the annual two-week session of the National People's Congress, or parliament, meeting in \textcolor{red}{Beijing} due to end on March 18.
				&
				 N angri textil ACTOR  ACTION-VERB  railway line in DIRECTIONAL ACTOR on DAY demand unpaid wage and unemplo ACTOR pay from  gover ACTOR  NONTOPIC
				 railway ACTOR saw ACTION-VERB.  N to N of  ACTION-VERB railway. stood on  railway later dispers  NONTOPIC  man  work at  railway station in sub-heilongjiang in  DIRECTIONAL  ADMIN of \textcolor{blue}{heilongjiang}.
				 train to \textcolor{red}{beijing} therefor delay N to N minut in leav NONTOPIC.  AVERB  similar to N in  MONZ N ACTAR  from   ACTAR upt textil plant ACTION-VERB  rail line  and   cut traffic on  airport highway  NONTOPIC ACTOR of embezzl  social secur pa ACTOR s.  ACTOR at \textcolor{blue}{sub-heilongjiang}  main train station NONTOPIC ADMINN  and  railway ACTOR went to persuad  to leav  and   ACTOR  dispers   N minut   NONTOPIC.  plight of disgruntl ACTOR  laid  from bloat state own firm  like  in \textcolor{blue}{sub-heilongjiang}   get top bill at  annual N week session of  nation peopl  ACTOR   par ACTOR  meet in \textcolor{red}{beijing} due to end on  MONTH  N.\ldots \\ \midrule
\multicolumn{2}{c}{ ICEWS location: \textcolor{red}{Beijing}---Correct location: \textcolor{blue}{Heilongjiang}  }\\
				\bottomrule
			\end{tabular}
	\end{minipage}
			\end{table}

Testing the context, sometimes called materiality, of the sentence that contains location words is another way of capturing relevant location words. The idea is that, if the sentence contains more action verbs and key nouns, the location word in that sentence is highly likely to be relevant. Likewise, a location word in a sentence with ``report'' or news agency names may be less likely to be relevant. 

Finally, we designed the data so that it can account for the variations at the article level as well as the data level, assuming that 1) some location words are more correct than others in each article and 2) some articles contain location words that are collectively more likely to be correct or incorrect altogether. In other words, these variables are calculated in relative terms within the article and data levels. This means that for the within article level variables, the location word with the largest value in that article has the value of 1. At the data level, only one location word with the largest value within the data has the value of 1. For example, the relative within article ratio of frequency for Heilongjiang in the example in Table \ref{tab:workingexample} would be 1 while that for Beijing would be 0.67. Table \ref{tab:workingexample} also shows the first and the second sentences containing ``heilongjiang'' include a irrelevant word ``said'', converted as ``NONTOPIC'', and therefore, the within-article materiality ratio for Heilongjiang and Beijing would be 0 and 0 while the immateriality ratio for the two would be 1 and 0 respectively. All the positive values would be much smaller in the within data ratios.    

To extract the above mentioned features, we start with the training set, which consists of two thirds of our text data. We first capture all location words that match the list in our location dictionary, and determine whether the location word falls into the correct category or the incorrect category, based on human coding. Once the recognized location word is determined as either correct or incorrect, they are stored separately for correct and incorrect corpora. 

Once the corpora of correct and incorrect N-grams are created, we compute the N-gram pattern information (N being the range specified\footnote{We computed this frequency rate for each location word for N-grams of two to seven.}), the frequencies of mention, and the materiality of the location sentences, in terms of both within article and data level ratios. The final data generated would contain the dependent variable indicating whether the particular location word is correct (Y=1) or not (Y=0), and the covariates of the frequencies of each N-gram and the three covariates mentioned above.

\begin{figure}
\includegraphics[width=6in]{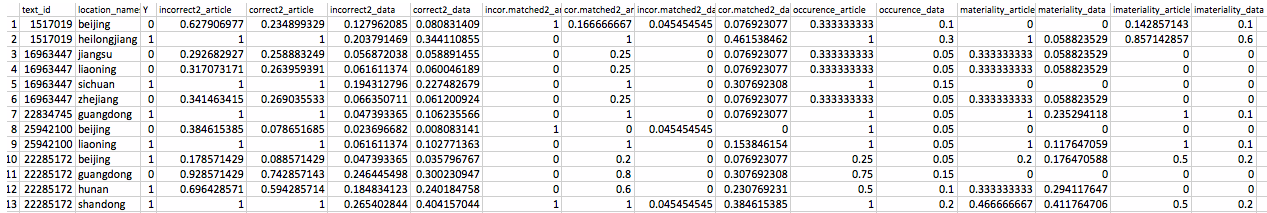}    
\caption{Data Generated (China)}\label{fig:data}
\end{figure}

Figure \ref{fig:data} shows the first thirteen rows and parts of covariates of the data generated using the Chinese news articles. The number of rows of the data equals the number of total province names appearing in all news articles. The first column represents the unique story IDs from ICEWS and the next column contains all of the location words in the article. The Y variable shows whether the location word is correct or not, based on the human coders' judgment. In the data shown, the first row represents the story 1517019 from the ICEWS data, the example in Table \ref{tab:workingexample}. The article contains two location words, `beijing' and `heiliongjian', of which the second is the correct event location. The next four location words in rows three through six are from a single article, ICEWS story 16963437. Of these, only `sichuan' is the correct event-relevant and event-occurring location.

The covariates with the suffix ``article'' are the relative ratios within articles. Location words within articles that do not contain any other locations are therefore assigned the value of one. The covariates with the suffix  ``data''  represent the relative ratios within the data. Correct and incorrect N-grams of two to seven, frequencies, and materiality variables are constructed in this manner.

In Stage 2, with the data (of the training set) generated, we fit the artificial neural networks, the support vector machine, and the random forests models. Using the random forests Recursive Feature Elimination (RFE) algorithm, we selected the variables of which the combination yields the highest accuracy rates in the training data.\footnote{The N-gram patterns were the most powerful variables consistently.} The parameters of each classifier were adjusted to get the optimal result. In random forests, the number of trees was set to 1000 and the kernel radial was set for SVM. The artificial neural networks model was tuned in each iteration, selecting automatically the best decay rate and the number of dendrites in the hidden layer. 

The estimated parameters were then used to predict the boolean status of each location word in the test set. For the example in Table \ref{tab:workingexample}, the average predicted probabilities for Beijing as the correct location is 6\% (neural net), 9\% (SVM), and 27\% (random forests) and those for Heilongjiang as the correct location is 97\% (neural net), 
75\%(SVM), and 98\% (random forests), making the predictions for \textit{both} location words correct. 

\section{Results}

Table \ref{table:validation} summarizes the performance statistics of various methods including our own classification approach. The columns represent each method and the rows indicate the data sets used. All numbers are rounded up. For the classification algorithms, we performed three 3-fold cross validations, thus the scores in the columns of neural net, SVM, and random forests are averages of nine iterations in total. The dictionary column represents the accuracy rate of the dictionary method that codes all captured location words as correct event locations.

\begin{table}[!ht]
\begin{center}
\caption{Validation Results}\label{table:validation}
\begin{tabular}{l|cccc}
& Classification  & Classification &  Classification & Dictionary \\
Datasets & (NNet) & (SVM) & (R.Forests) &  \\
\hline
China (ICEWS, Protest) & 73 & 75 &  73 & 51  \\
D. R. Congo (ICEWS, Fight) & 81 & 84 & 84 & 59 \\
Syria (OEDA, Fight) & 74 & 73 & 75 & 57\\
\end{tabular}

\end{center}
\end{table}

\begin{figure}[!ht] 
\centering
    \includegraphics[width=6in]{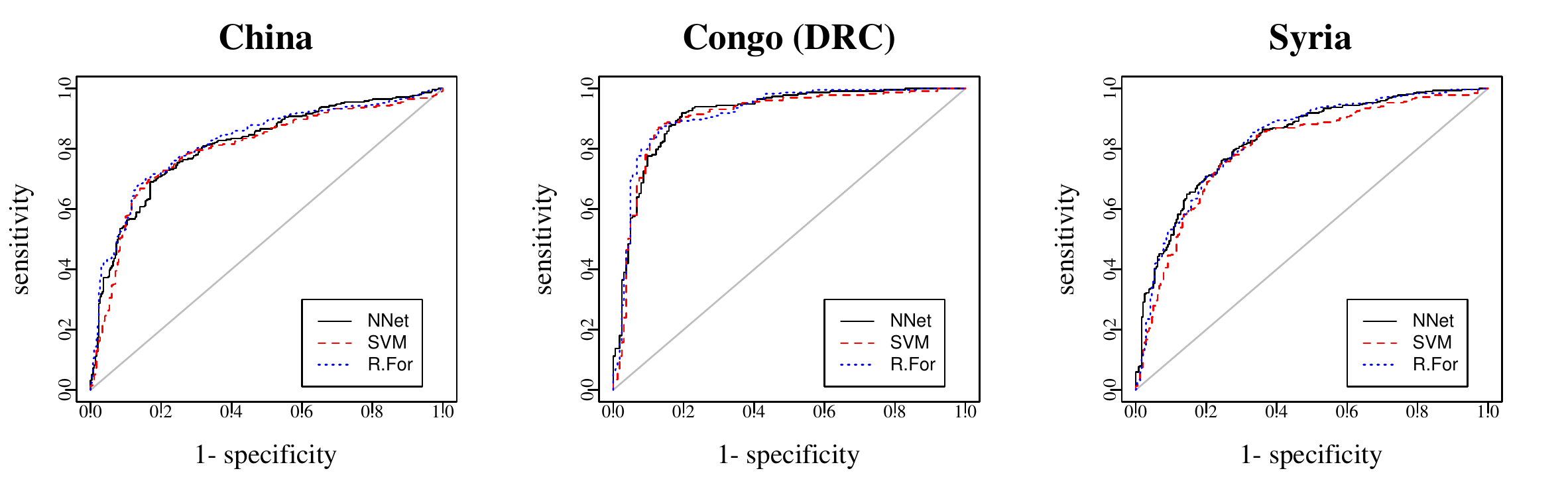}
    \caption{Receiver operating characteristic Curves}\label{fig:ROC}
\end{figure}

As Figure \ref{fig:ROC} shows, the accuracy rates across models do not vary much with about 3\% maximum difference. While the performance of our algorithm is consistently high across classifiers, the highest accuracy rates in each data set were produced by SVM for the China and DRC data, and by random forests for the Syria data. However, the results vary across data sets, with the highest rates in the DRC data. This difference comes from the number of true locations in the article. The accuracy rates for location words in the subset consisting of only articles with one correct location  range from 86\% in China to 90\% in the DRC. 

\begin{figure}[!t] 
\includegraphics[width=2in]{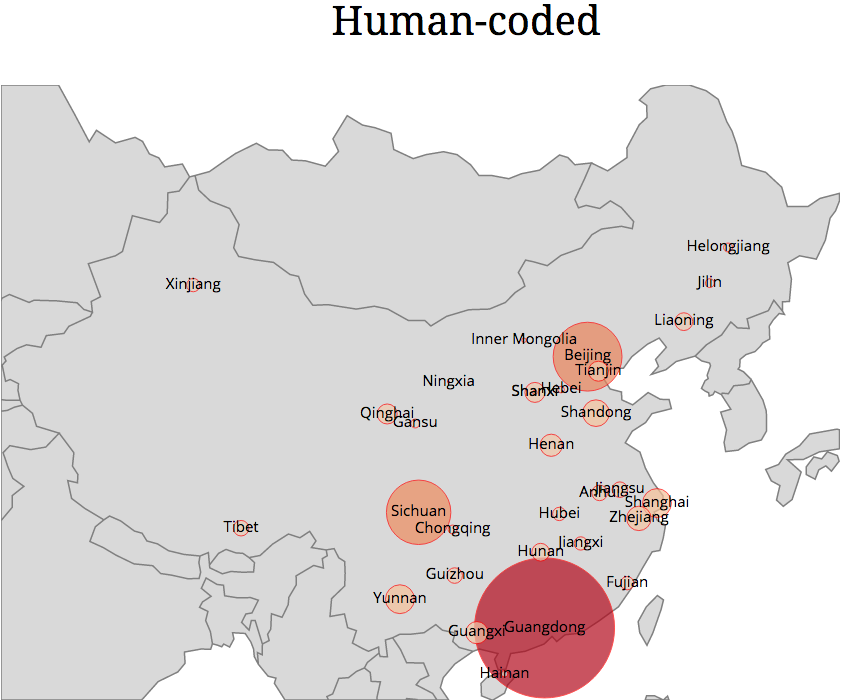}\includegraphics[width=2in]{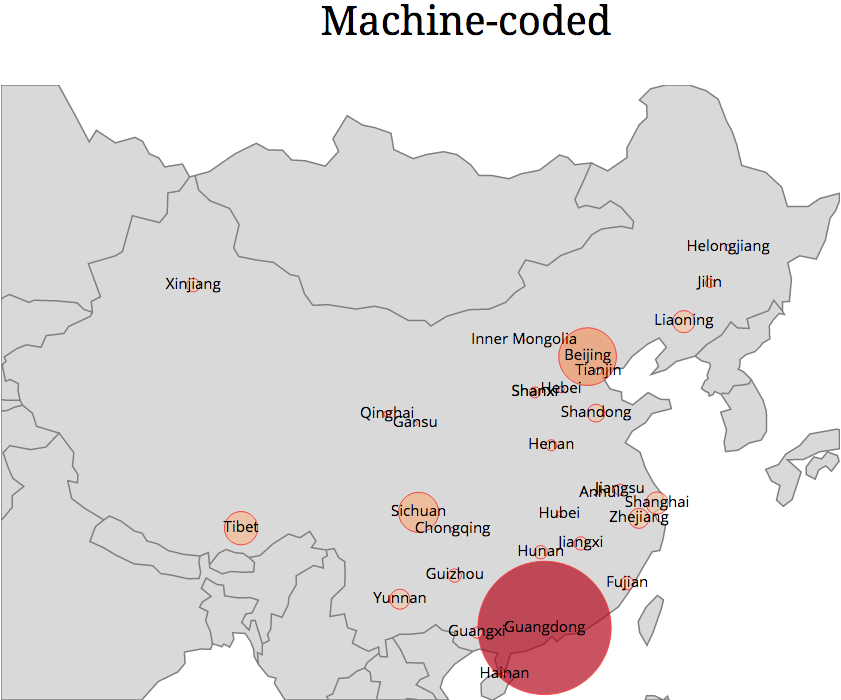}\includegraphics[width=2in]{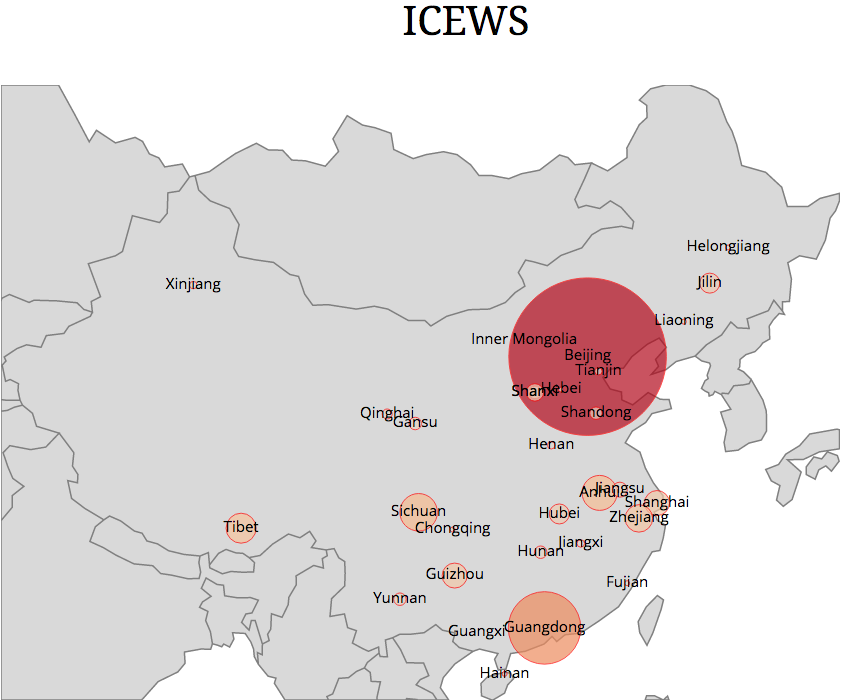}\includegraphics[width=0.3in]{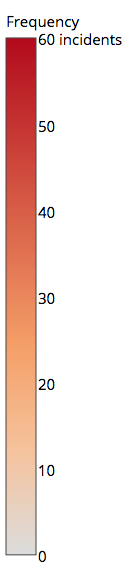}
\caption{Protest Frequencies in China}\label{fig:bubblechina}
\end{figure}

We have compared the true positives of our results to the current machine-coded data. These are the location words that each algorithm classifies as the actual event locations. Figures \ref{fig:bubblechina} to 7 compare the results to the ground truth which is plotted on the left and the current ICEWS or OEDA locations on the right. The sizes of bubbles represent the relative shares of events and the colors represent frequencies with the legends on the far right. In the China data, our algorithm misses eleven protests in Sichuan, but in all other provinces the differences are single digits. On the other hand, ICEWS codes Beijing as the event location more than 30 cases than the ground truth and misses more than 30 protest cases in Guangdong province alone.

\begin{figure} 
\includegraphics[width=2in]{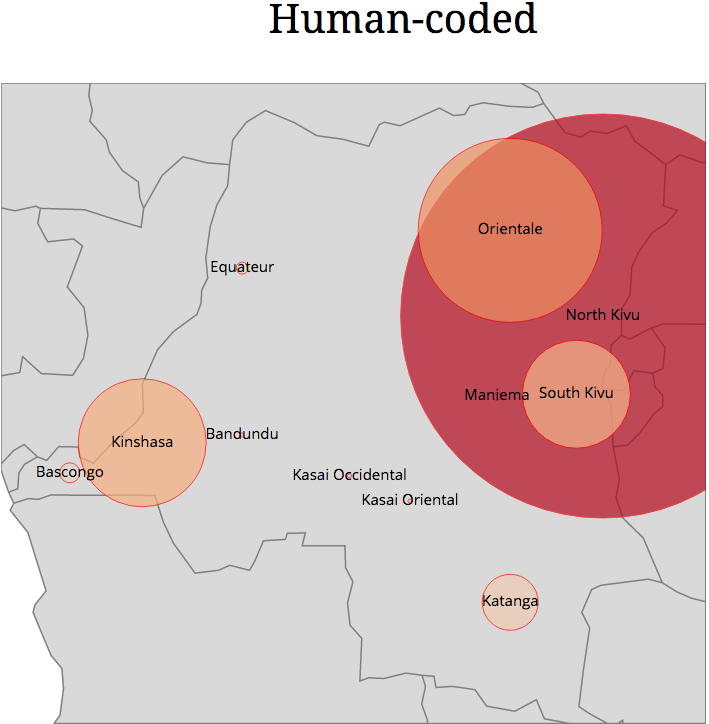}
\includegraphics[width=2in]{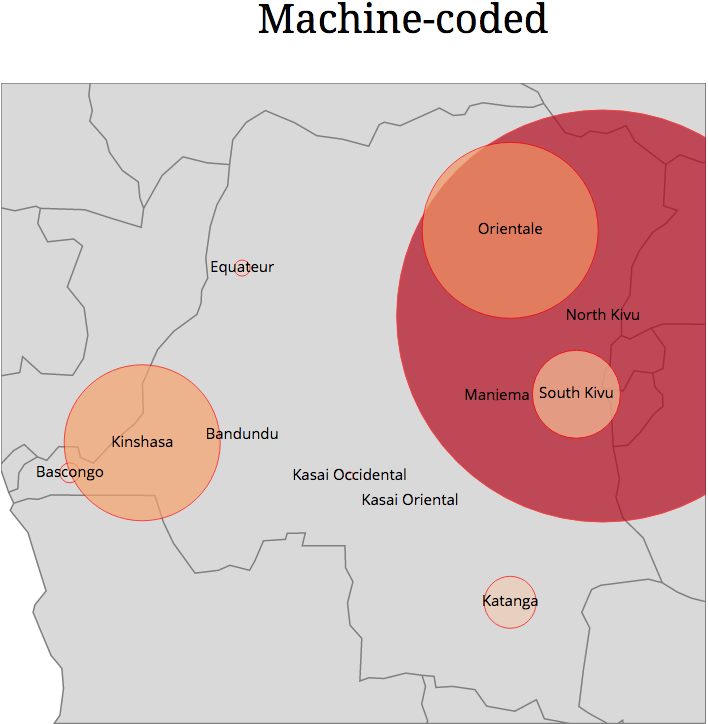}
\includegraphics[width=2in]{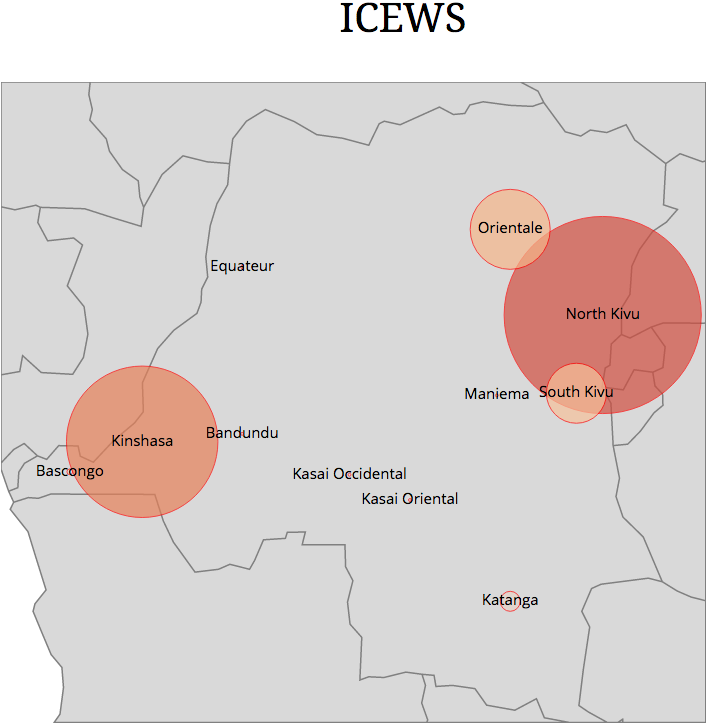}
\includegraphics[width=0.3in]{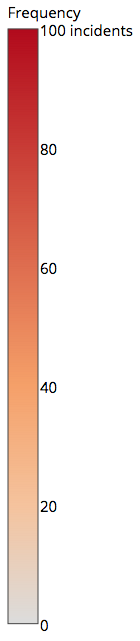}
\caption{Fight Frequencies in Democratic Republic of Congo
\label{fig:bubbleDRC}
}
\end{figure}

Results in the DRC are accurate regardless of the choice of classifiers and the best performance is around 85\%. This is true in part because the events in the DRC do not typically include a large number of locations. The civil conflicts which show up in the fight category in the DRC are concentrated in a small number of areas.  By comparision, protests in China are not only in a much larger country, but are in a wide variety of locations. Accordingly, stories about China have many more location words per story, and are harder to correctly identify than is the case in the DRC.  Figure \ref{fig:bubbleDRC} shows that the bubbles of the human-coded map on the left and the machine-coded map in the middle are nearly identical. Compared to the human coded locations, the locations coded by ICEWS model are correct at around 67\% with over 30 under-reporting cases of fight in North Kivu and over 20 over-reporting cases in Kinshasa.

In the Syria case, our overall predicted event locations also look very similar to the ground truth while OEDA not only misses  over one-half of the true event locations  (129 NAs in 250 news articles), but also includes event locations that are not in Syria such as Beirut (three events), Illinois (one event), Moscow (one event), New Jersey (one event) and Pennsylvania (four events). Compared to the human coded locations, event locations in OEDA data 
are correct 31\% of the time. 

Overall, the performance of our classifiers is strong, improving the accuracy rate by as much as 25\% from the dictionary approach. Furthermore, even if the accuracy rate is not 100\%, because our algorithm evaluates each location word, it does not symmetrically miss or favor certain locations as the current ICEWS and OEDA algorithms do. Hence, the visualized results seem very close to the ground truth. 

\begin{figure} 
\includegraphics[width=2in]{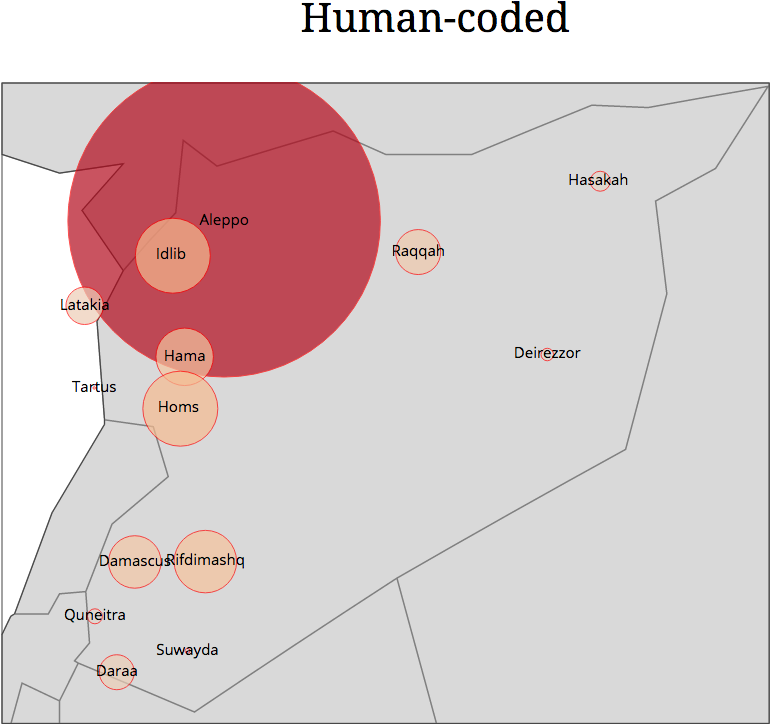}\includegraphics[width=2in]{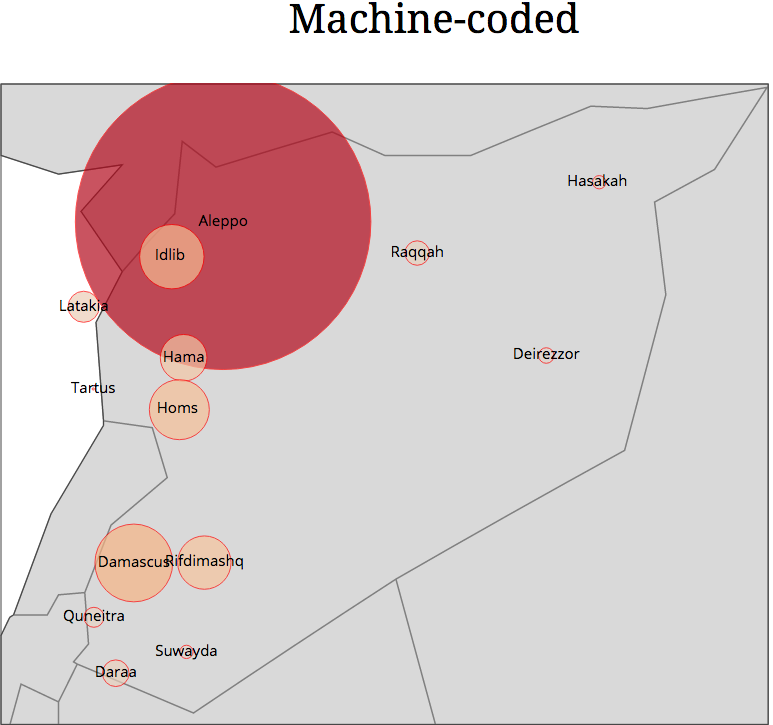}\includegraphics[width=2in]{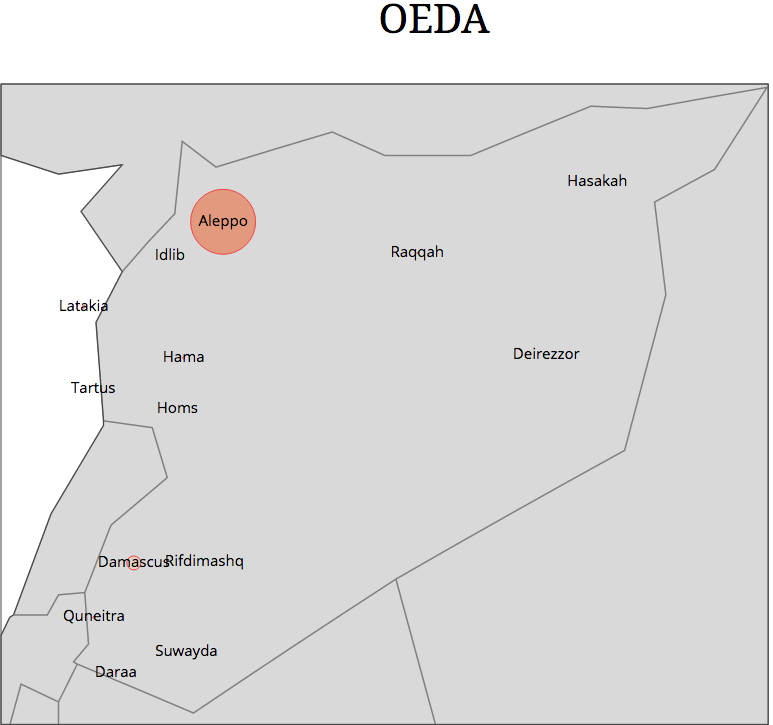}\includegraphics[width=0.3in]{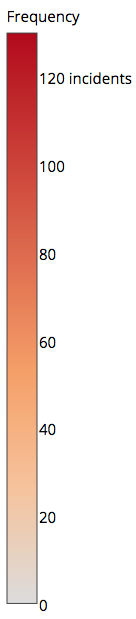}
\caption{Fight Frequencies in Syria}\label{fig:bubblesyria}
\end{figure}

\section{Conclusion} 

We examined some problems associated with current geolocation methods employed in existing machine-coded event data. Locations of events contain valuable information that is of interest to many scholars and policy makers.
To address discrepancies in geolocation  between automated and human coders, we developed a supervised machine learning algorithm that filters out event-irrelevant locations as well as non-event-occurring ones. Departing from the assumption that one correct location exists in a news article, we evaluate each location word. By doing so, we diverge from algorithms that are systematically biased towards certain locations such as the capital of a country and locations that appear frequently in the corpus. Using human coded ground-truth, we demonstrate that this approach is superior to extant approaches in the cases we have studied. 

Interested scholars can extend the  current work to a wider range of event ontologies and locations. While we have studied only a few countries, the protocol we developed may aid others who are interested in different countries to geolocate extant event data more accurately. Others who wish to extract location information from structured text data written in formal language, such as the United Nations reports on Children and Armed Conflict (\texttt{https:childrenandarmedconflict.un.org}) or Amnesty International country reports (\texttt{https:www.amnesty.orgen/latest/research/2016/02/annual-report-201516/}), can utilize our (open source) protocol---available upon publication---to create new event data streams in which the events are geolocated.


\bibliography{/Users/mdw/Git/whistle/master}
\bibliographystyle{authordate1}

\newpage

\end{document}